\documentclass{article}

\usepackage[preprint]{corl_2026} 
\usepackage{booktabs}
\usepackage{multirow}
\usepackage{graphicx}
\setkeys{Gin}{draft=false}
\usepackage{caption}
\usepackage{amsmath}
\usepackage{amssymb}
\usepackage{wrapfig}
\usepackage{enumitem}

\title{Physics-Aware Object-Centric SE(3) Forecasting for Real-Time Proactive Dynamic Grasping}

\author{
\begin{tabular}{ccccc}
Yirui Zhou\textsuperscript{1,*} &
Hongli Xu\textsuperscript{2,*} &
Ying Zheng\textsuperscript{1,$\dagger$} &
Yi Wang\textsuperscript{1,$\dagger$} &
Lap-Pui Chau\textsuperscript{1,$\dagger$}
\end{tabular}
\\[2mm]
\textsuperscript{1}The Hong Kong Polytechnic University
\textsuperscript{2}Technical University of Munich
\\[1mm]
\textsuperscript{*}Equal contribution.
\qquad
\textsuperscript{$\dagger$}Corresponding authors.
}

\begin{document}
\maketitle

\hypersetup{
  pdftitle={Physics-Aware Object-Centric SE(3) Forecasting for Real-Time Proactive Dynamic Grasping},
  pdfauthor={Yirui Zhou, Hongli Xu, Ying Zheng, Yi Wang, Lap-Pui Chau}
}

\begin{figure}[!ht]
    \vspace{-1.8em}
    \centering
    \includegraphics[width=1.0\linewidth]{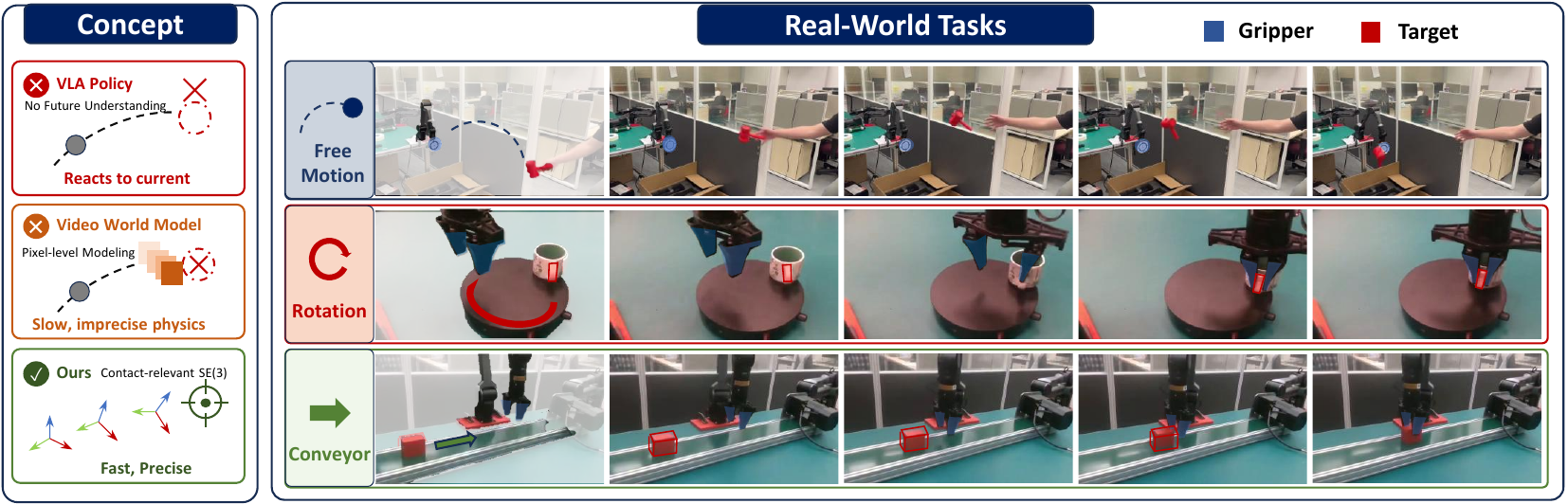}
    \caption{Left: Reactive policies act on the current object pose and fail under fast dynamics. Video world models predict pixel-level futures that are slow and physically imprecise. Our method predicts contact-relevant SE(3) state — compact, fast, and action-guided. Right: Real-world dynamic grasping tasks including parabolic catching, conveyor interception, and turntable grasping.}
    \label{fig:teaser}
    \vspace{-1.0em}
\end{figure}

\begin{abstract}
Proactive dynamic grasping requires robots to predict the future state at which physical contact will occur, rather than reacting to the currently observed object pose. But what future must the robot predict, and at what level of abstraction? We argue that contact-relevant rigid-body state in SE(3) is a compact task-sufficient abstraction for this problem: it directly determines where and when the robot should establish contact, without requiring visually complete future scenes. This insight leads to a lightweight object-centric forecast-then-select framework that predicts future rigid-body trajectories from short pose histories and uses a reachability-aware interception selector to choose feasible contact states under kinematic and timing constraints. Trained with randomized physics simulation over primitive shapes, the forecaster transfers to real objects once their metric geometry and SE(3) state histories are recovered. Experiments across simulation and real-world dynamic grasping tasks show that our framework improves the average success rate from \(24.5\%\) to \(62.0\%\) in simulation, and from \(15.3\%\) to \(67.3\%\) in real-world catching, conveyor, and turntable interception. The SE(3) forecaster runs in \(15\,\mathrm{ms}\), orders of magnitude faster than dense visual future prediction, enabling real-time contact decision making. Our results suggest that effective proactive grasping does not require imagining the entire future scene. Instead, predicting the compact physical future that determines contact provides an effective and practical abstraction for real-time dynamic grasping. Project page: \url{https://am0rarg0.github.io/paf-grasp/}
\end{abstract}


\section{Introduction}
Humans grasp moving objects not by reacting to the current state, but by anticipating where they will be at contact time. Catching a tossed key, intercepting a sliding snack, or grabbing a swinging handle require predicting where contact will occur and acting proactively. Robotic dynamic grasping faces the same challenge, raising a fundamental question: what future should the robot predict, and at what level of abstraction? As illustrated in Fig.~\ref{fig:teaser}, we argue that the answer to this question determines whether proactive grasping in dynamic scenarios can be reliable, efficient, and physically grounded.

Existing work approaches robot manipulation from two broad directions: reactive policy and predictive world modeling. Visuomotor policies and vision-language-action (VLA) models have made substantial progress by mapping visual observations and language instructions directly to robot actions~\citep{act, chi2025diffusion, ze20243ddiffusionpolicygeneralizable, act3d, xie2026dynamicvla, black2026pi0visionlanguageactionflowmodel, intelligence2025pi05visionlanguageactionmodelopenworld, intelligence2025pi06vlalearnsexperience, Tang2025VLASHRV, nvidia2025gr00tn1openfoundation, meng2025preserving, yirui}. While temporal observations, action chunking, and real-time policy updates improve responsiveness, these policies are commonly executed as reactive observation-to-action systems. For dynamic grasping, the key decision target is the future object state at which physical contact becomes feasible. This contact-time state depends on object motion, robot reachability, and system latency, but is not typically made the explicit decision variable in end-to-end policy
execution. A parallel line of work instead studies future prediction through world models and video generation, anticipating the future by generating future observations or videos~\citep{zhang2026foreact, kim2026cosmos}, predicting latent visual features~\citep{li2026causalworldmodelingrobot}, or modeling joint video-action trajectories~\citep{ye2026worldactionmodelszeroshot}. While such models provide powerful visual foresight, pixel-level prediction is
an indirect and computationally heavy proxy for contact-time reasoning. Dynamic
grasping requires metrically accurate future object pose and motion rather than
visually complete future scenes, and video-level rollouts introduce computational
overhead that is misaligned with the real-time demands of dynamic interception.


In this work, we argue that the question for proactive dynamic grasping
is not whether to predict the future, but what future to predict and at what
level of abstraction. We identify contact-relevant rigid-body state in SE(3) as a compact and actionable abstraction: it is aligned with rigid-body dynamics, metrically precise, and directly determines where and when contact should occur without requiring complete future scenes. We instantiate this idea as \textbf{PAF-Grasp}, a physics-aware forecast-then-select framework that couples an object-centric continuous-time forecaster with a reachability-aware interception selector. While prior motion-aware dynamic manipulation methods use task-specific future cues or reactive execution~\citep{li2026simtorealdynamicobjectmanipulation, DBC-TFP}, PAF-Grasp explicitly models the future state by forecasting rigid-body motion from pose histories and selecting feasible contact states under kinematic and timing constraints~\citep{metacontroller, reacha}. This separates object-centric prediction from robot-specific execution and converts dynamic grasping from reactive pose chasing into proactive contact-state selection.

We evaluate PAF-Grasp on three design questions: the value of explicit
contact-state prediction, SE(3) future representation, and the need for
real-time contact decisions. Explicit contact-state prediction improves
success rate by \(37.5\) and \(52.0\) percentage points in simulation and
real-world catching, conveyor and turntable interception, while achieving higher contact-relevant accuracy than dense visual world models at orders-of-magnitude lower latency. These results show that an action-relevant abstraction is critical for proactive dynamic grasping. Our contributions are threefold:
\begin{itemize}[leftmargin=*, itemsep=0pt, topsep=1pt, parsep=0pt, partopsep=0pt]
    \item We identify contact-relevant SE(3) rigid-body state as a compact abstraction for proactive dynamic grasping, and provide systematic evidence that physical state prediction outperforms dense visual future prediction
    alternatives in accuracy, physical consistency, and real-time efficiency.
    \item We propose a lightweight SE(3) object-centric forecaster that predicts future rigid-body trajectories from short pose histories, coupled with a reachability-aware interception selector for feasible contact-state selection under kinematic and timing constraints.
    \item We validate PAF-Grasp across simulated and real-world dynamic grasping tasks, demonstrating higher success rates than reactive visuomotor baselines while maintaining real-time efficiency.
\end{itemize}

\section{Related works}
\label{sec:citations}
\subsection{Dynamic Objects Grasping}
Visuomotor policies have shown strong capabilities for robot manipulation in static or quasi-static environments
~\citep{black2026pi0visionlanguageactionflowmodel,
intelligence2025pi06vlalearnsexperience,chi2025diffusion,
ze20243ddiffusionpolicygeneralizable}. Extending these policies to dynamic
object manipulation requires the robot to act under perception latency, control
delay, and rapidly changing object states \citep{zheng2025survey}.

Prior work on dynamic grasping, detection, and tracking
~\citep{fang2023anygrasp, MotionGrasp, TrreactiveGrasping}
has mainly studied how to maintain stable object or grasp-pose estimates over
time, enabling reactive execution as the target moves. Other approaches have
introduced motion-aware planning or future pose prediction for dynamic
manipulation~\citep{metacontroller, DBC-TFP, reacha}, typically focusing on settings where object
motion follows structured, repeatable, or task-specific patterns. Recent VLA-based systems further improve responsiveness in dynamic scenes.
For example, DOMINO~\citep{li2026simtorealdynamicobjectmanipulation} has advanced this line
of work by introducing a large-scale benchmark for dynamic grasping and
showing that historical motion cues and auxiliary future-feature prediction can
improve dynamic-aware VLA policies. DynamicVLA~\citep{xie2026dynamicvla}
further improves responsiveness through real-time action chunking. These works indicate a broader shift from static policy learning toward
temporally aware and motion-aware manipulation systems. However, dynamic grasping across
projectile, linear, rotational, and sliding motions requires explicit
contact-time prediction under reachability and latency constraints. We therefore
study object-centric SE(3) trajectory forecasting as a decision representation
for proactive dynamic grasping.

\subsection{Future Prediction via World Models}
World models have become an increasingly popular paradigm for future prediction
in embodied intelligence. One line of work predicts future observations with
large-scale video generation models~\citep{wan2025wanopenadvancedlargescale,
yangcogvideox, klingteam2025klingomnitechnicalreport, hu2023gaia},
producing visually realistic videos from text, image, or action-conditioned
inputs. A related line integrates future prediction with policy learning through
world-action models~\citep{ye2026worldactionmodelszeroshot,fastwam,
motubrainteam2026motubrainadvancedworldaction, li2026causalworldmodelingrobot}, where video modeling provides a
dense representation of how the world evolves and can improve policy
generalization. However, dynamic grasping is a contact-relevant and latency-sensitive setting: objects can move substantially within a short window, making
pixel-level rollouts costly, indirect, and physically unreliable under fast
dynamics.

Another line of work explores latent world models that predict futures in representation spaces rather than pixel space. Latent imagination
models learn predictive dynamics for planning, while Joint Embedding Predictive Architecture (JEPA)-like models~\citep{Assran_2023_CVPR,bardesrevisiting} predict abstract future
representations without reconstructing pixels. Semantic latent diffusion world
models further show that pretrained representations can preserve
action-relevant structure and task progress beyond reconstruction fidelity
alone~\citep{jha2026reconstruction}. Physically structured world-model
perspectives emphasize physically grounded latent dynamics rather than
unconstrained visual sequence prediction~\citep{cui2026physically}. Our work takes a different route: rather than learning a general-purpose latent
world model, we make the predictive state explicit as object-centric SE(3)
rigid-body state. This makes the predicted future directly usable for
contact-time reachability and interception.
\section{Method}
\begin{figure}[t]
    \centering
    \includegraphics[width=1.0\linewidth]{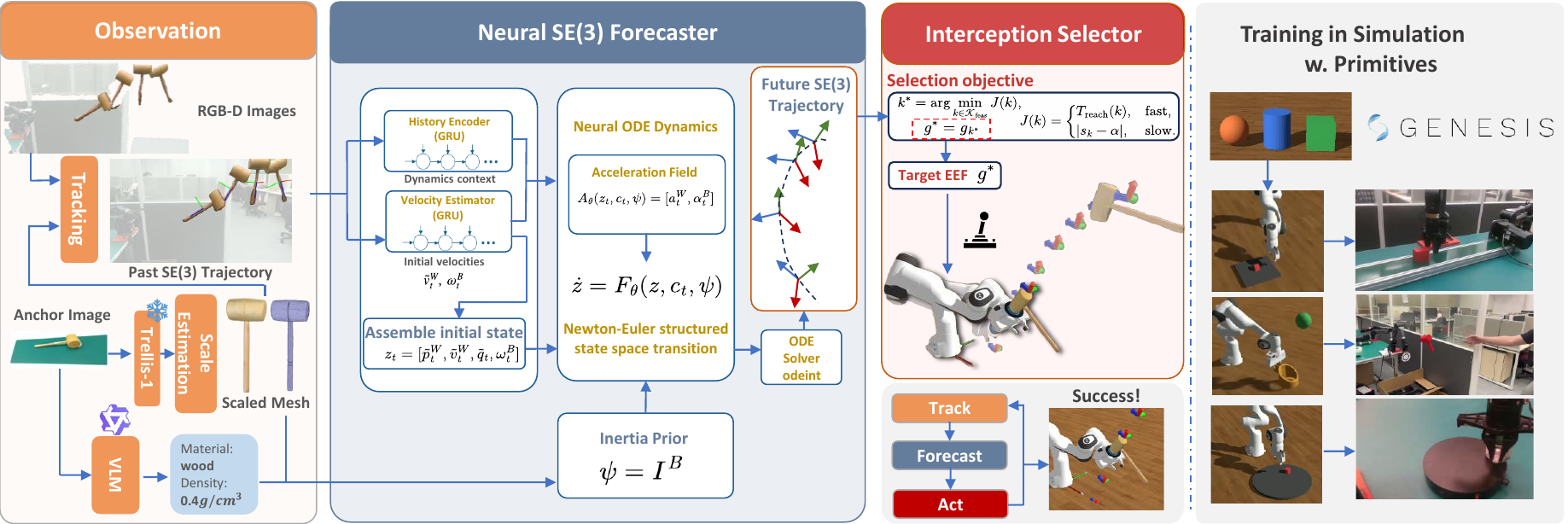}
    \caption{
    Overview of PAF-Grasp. The Neural SE(3) forecaster is trained on
    Genesis-generated primitive-object trajectories to learn reusable rigid-body dynamics. At deployment, an anchor RGB-D observation is used offline to recover object-specific geometry and an approximate inertia prior, while the online RGB-D stream continuously updates the object's SE(3) history for closed-loop
forecasting and interception. The forecaster encodes recent motion history, estimates initial velocities, and integrates a Newton--Euler-structured neural ODE to predict future rigid-body trajectories. The forecast-guided interception selector then propagates task-specific contact frames along the predicted trajectory and selects a reachable end-effector target under timing and kinematic constraints.}
    \label{fig:main}
    \vspace{-1.2em}
\end{figure}

Fig.~\ref{fig:main} overviews PAF-Grasp: a sim-to-real forecast-then-select framework with a physics-aware SE(3) forecaster and a reachability-aware interception selector for proactive dynamic grasping.
\subsection{Problem Formulation}

We study proactive grasping of a moving rigid object. At time \(t\), the
object-to-world pose is represented as \(T_t^W=(R_t^W,p_t^W)\in\mathrm{SE}(3)\), where \(R_t^W\in\mathrm{SO}(3)\) and
\(p_t^W\in\mathbb{R}^3\) denote the object's orientation and position in the world frame. Given a pose history \(\mathcal{H}_t=\{T_{t-K+1}^W,\ldots,T_t^W\}\), where \(K\)
is the observation horizon, the forecaster predicts future object poses
\(\hat{\mathcal{Y}}_t=\{\hat T_{t+1}^W,\ldots,\hat T_{t+H}^W\}\) over a
prediction horizon \(H\). Dynamic grasping then reduces to selecting a contact-time state from this predicted trajectory. For each task, we define an object-frame attachment transform
\(A^o\in\mathrm{SE}(3)\) on the reconstructed geometry, obtained from task
teaching, geometric rules, or grasp proposals, and representing a grasp frame,
catching point, or functional tool frame. Given a predicted object pose
\(\hat T_{t+k}^W\), this induces the future task frame
\(\hat T_{t+k}^{A}=\hat T_{t+k}^W A^o\). A task-dependent mapping \(\Phi\)
aligns the robot-side interaction frame with this task frame using a fixed
contact-to-EEF offset, yielding \(g_k=\Phi(\hat T_{t+k}^{A})\).
The robot selects the final target \(g^*=g_{k^*}\) from feasible future
candidates:
\begin{equation}
    k^*=\arg\min_{k\in\mathcal{K}_{\mathrm{feas}}} J(k),
    \qquad
    g^*=g_{k^*},
\end{equation}
where \(J(k)\) is a task-dependent interception cost and
\(\mathcal{K}_{\mathrm{feas}}\) contains candidates satisfying reachability and
timing constraints. This separates object-centric forecasting from task-specific execution.

\subsection{Object-Centric Dynamic Forecaster}
We design an object-centric forecaster that predicts future rigid-body motion
from short pose histories. For each trajectory window, we represent motion
relative to the last observed object pose using mixed coordinates: translation
is the world-frame displacement from the last observed position, while rotation
is relative to the last observed object orientation. We formulate second-order rigid-body dynamics as a first-order ODE over the augmented pose-velocity state
\(z_t=[\tilde p_t^W,\tilde v_t^W,\tilde q_t,\omega_t^B]\), where
\(\tilde p_t^W\) and \(\tilde v_t^W\) are relative position and velocity in the
world frame, \(\tilde q_t\) is the relative rotation, and \(\omega_t^B\) is the
body-frame angular velocity. At rollout start, \(\tilde p_t^W=0\) and
\(\tilde q_t\) is the identity rotation. After rollout, predictions are
converted back to world-frame poses by
\(\hat p_{t+k}^W=p_t^W+\hat{\tilde p}_{t+k}^W\) and
\(\hat q_{t+k}^{WB}=q_t^{WB}\otimes\hat{\tilde q}_{t+k}\).
The recovered poses \(\hat T_{t+k}^W\) are used for contact-state selection.
This object-centric representation removes global workspace bias while
preserving body-frame rotational structure. We use Newton--Euler dynamics as a structural prior in the same mixed
coordinates, with translation in the world frame and rotation in the body frame:
\begin{equation}
\label{eq:newton_euler}
\left\{
\begin{aligned}
    \dot{\tilde p}^{W} &= \tilde v^{W}, \\
    \dot{\tilde q} &= \frac{1}{2}\tilde q\otimes[0,\omega^B], \\
    m\dot{\tilde v}^{W} &= f^{W}, \\
    I^B\dot\omega^B &=
    \tau^B-\omega^B\times(I^B\omega^B).
\end{aligned}
\right.
\end{equation}
Here, \(m\) and \(I^B\) denote object mass and body-frame inertia, while
\(f^W\) and \(\tau^B\) denote net force and torque. We do not identify these
external inputs explicitly. Instead, we learn the induced accelerations
\(a^W=\dot{\tilde v}^{W}\) and \(\alpha^B=\dot\omega^B\), conditioned on an
approximate inertia prior \(\psi=I^B\).

Since pose observations provide noisy finite-difference velocities, we estimate
the initial translational and angular velocities with a lightweight gated recurrent unit (GRU)-based velocity estimator. Another GRU encodes recent pose-motion history into a
dynamics context \(c_t\). Given \(z_t\), \(c_t\), and \(\psi\), an MLP predicts
the acceleration field
\(A_\theta(z_t,c_t,\psi)=[a_t^W,\alpha_t^B]\), where \(a_t^W\) is in the world
frame and \(\alpha_t^B\) in the body frame. This induces the continuous-time
state-space transition
\begin{equation}
\label{eq:state_space_transition}
    \dot z
    =
    F_\theta(z,c_t,\psi)
    =
    \left[
    \tilde v^W,\;
    a^W,\;
    \frac{1}{2}\tilde q\otimes[0,\omega^B],\;
    \alpha^B
    \right],
    \quad
    [a^W,\alpha^B]=A_\theta(z,c_t,\psi),
\end{equation}
which is integrated over \(\{0,\Delta t,\ldots,H\Delta t\}\) from \(z_t\) to
obtain \(\hat z_{t:t+H}\). We extract poses from the rollout
and apply quaternion normalization and sign alignment during integration. This continuous-time
formulation enables prediction at arbitrary future times and improves stability
under horizon shift compared with recursive autoregressive rollout. The model is
optimized with
\(\mathcal{L}
= \lambda_{\mathrm{pos}}\mathcal{L}_{\mathrm{pos}}
+ \lambda_{\mathrm{rot}}\mathcal{L}_{\mathrm{rot}}
+ \lambda_{\mathrm{vel}}\mathcal{L}_{\mathrm{vel}}\),
where \(\mathcal{L}_{\mathrm{pos}}\) and \(\mathcal{L}_{\mathrm{vel}}\) are MSE
losses over translation and velocity, and \(\mathcal{L}_{\mathrm{rot}}\) is a
sign-invariant quaternion chordal loss. The fixed loss weights are provided in
Appendix~\ref{app:implementation}.

\subsection{Sim-to-real Rigid-body State Forecasting}
The forecaster is trained on Genesis~\citep{Genesis} generated rigid-body
trajectories using primitive shapes, such as spheres, cuboids, and cylinders,
with domain randomization over initial conditions and physical parameters. For
each simulated primitive, its metric geometry and density provide the
corresponding body-frame inertia prior \(\psi=I^B\). We choose realistic ranges for release velocity, release height, friction,
material density, and ramp angles using common physical ranges and LLM~\citep{singh2026openaigpt5card}-assisted priors. Rather than matching the exact geometry of real test objects, these
trajectories provide supervision for rigid-body motion. This transfer is enabled by the dynamics abstraction in
Eq.~\eqref{eq:newton_euler}. Unlike autoregressive forecasters that roll out
future poses through discrete kinematic transitions, our model represents object
motion as continuous-time dynamics with a second-order state and a learned
acceleration field. The inertia prior provides object-specific rotational
structure, while the learned acceleration field absorbs residual effects from smooth unmodeled forces, friction, constraints, density error, and perception noise within the training regimes. This encourages transferable motion evolution beyond short-horizon pose transitions of specific training instances. We use the same physics-structured forecaster architecture across all motion families, while training one checkpoint per motion family. Each checkpoint is trained on randomized primitive objects and transfers across object instances within that motion family without object-specific retraining.

For real-world deployment, we separate offline object initialization
from online closed-loop execution. Before execution, a single anchor
RGB-D observation is used to recover a metric object mesh and an
approximate body-frame inertia prior \(I^B\) through segmentation,
mesh reconstruction, render-based metric alignment, and coarse
material-density estimation~\citep{Ravi2024SAM2S,trellis,gengone,bai2025qwen3}.
The resulting geometry and \(\psi=I^B\) remain fixed during execution
and enable the corresponding motion-family forecaster to transfer to
real objects without object-specific retraining. Online, an RGB-D
stream continuously updates the SE(3) pose history, from which the
velocity estimate, dynamics context, future trajectory, and interception
target are recomputed at 10\,Hz until contact. The forecaster runs in
15\,ms. Details of the sim-to-real pipeline are provided in
Appendix~\ref{app:sim2real}.

\subsection{Forecast-guided Interception Selector}
Following classical robotic interception formulations~\cite{rendezvous,lampariello2011catching},
our selector chooses a feasible future rendezvous frame along the predicted
target trajectory instead of reacting to the current state. Given
\(\{\hat T_{t+k}^{A}\}_{k=1}^{H}\), each candidate time
\(\tau_k=k\Delta t\) is mapped to a desired end-effector contact pose
\(g_k=\Phi(\hat T_{t+k}^{A})=(p_k^g,q_k^g)\). We first estimate the time required
to reach each candidate and keep only safe candidates that are reachable before
the predicted contact time:
\begin{equation}
\label{eq:feasible_set}
\begin{aligned}
T_{\mathrm{reach}}(k)
&=
\max\!\left(
\frac{\|p_k^g-p_t^e\|}{\bar v_e},
\frac{d_R(q_k^g,q_t^e)}{\bar\omega_e}
\right)
+
T_{\mathrm{lat}}, \\
\mathcal{K}_{\mathrm{feas}}
&=
\left\{
k:
g_k\in\mathcal{W}_{\mathrm{safe}},
\;
T_{\mathrm{reach}}(k)+T_{\mathrm{margin}}\le\tau_k
\right\}.
\end{aligned}
\end{equation}
Here, \(d_R\) is the quaternion geodesic distance, \(T_{\mathrm{lat}}\)
accounts for perception, inference, communication, and controller latency,
\(T_{\mathrm{margin}}\) handles timing uncertainty, and
\(\mathcal{W}_{\mathrm{safe}}\) encodes workspace, IK, height, and collision
constraints. The reach-time estimate jointly accounts for translational and
rotational motion. After removing candidates that cannot be reached in time,
fast tasks choose the feasible target with the smallest estimated reach time,
while slow tasks choose a desired relative contact time within the feasible
window:
\begin{equation}
\label{eq:interception_selection}
k^*=\arg\min_{k\in\mathcal{K}_{\mathrm{feas}}}J(k),
\qquad
J(k)=
\begin{cases}
T_{\mathrm{reach}}(k), & \mathrm{fast},\\[2pt]
|s_k-\alpha|, & \mathrm{slow}.
\end{cases}
\end{equation}
For slow tasks, we sort
\(\mathcal{K}_{\mathrm{feas}}=\{k_i\}_{i=1}^{M}\) by contact time
\(\tau_{k_1}\leq\cdots\leq\tau_{k_M}\), and set
\(s_{k_i}=(i-1)/(M-1+\epsilon)\). The term \(|s_k-\alpha|\) is the normalized
rank error to the desired contact-time position \(\alpha\in[0,1]\), from earliest
\((0)\) to latest \((1)\).

\section{Experiments}
\label{sec:result}
Our experiments evaluate whether contact-time rigid-body prediction enables
proactive dynamic grasping. We focus on three questions:
\textbf{(i)} whether future contact-state prediction improves dynamic grasping;
\textbf{(ii)} which future representation best supports real-time interception;
and \textbf{(iii)} how forecasting accuracy interacts with kinematic feasibility.
We study these questions across simulated and real-world dynamic grasping tasks. Protocol details, hardware setup, motion ranges,
randomization, and control-frequency settings are provided
in Appendices~\ref{app:sim_tasks}, \ref{app:real_tasks},
and~\ref{app:baselines}.

\subsection{Overall Performance on Dynamic Grasping Tasks}
\textbf{Simulation Experiments.}
We evaluate on four dynamic grasping tasks: parabolic catching, conveyor,
turntable, and ramp sliding interception. We compare against
reactive visuomotor policies, including
\(\pi_0\)~\citep{black2026pi0visionlanguageactionflowmodel},
\(\pi_{0.5}\)~\citep{intelligence2025pi05visionlanguageactionmodelopenworld},
\(\pi_0\)-FAST~\citep{pertsch2025fastefficientactiontokenization},
VLASH~\citep{Tang2025VLASHRV}, and GR00T N1.7~\citep{nvidia2025gr00tn1openfoundation},
all fine-tuned and evaluated under the same Genesis-generated data,
initialization, and success criteria. OURS uses the same interface as
\(\pi_{0.5}\), but augments it with explicit SE(3) forecasting and
reachability-aware contact-state steering~\citep{christopher2024constrained,
yao2025inference}. This comparison evaluates the effect of making a reachable
future contact state explicit, rather than equalizing intermediate
representations across methods. All methods use the same task data,
initialization, and success criteria. We use ground-truth object poses as
forecaster inputs and evaluate 100 trials for Catch and 50 trials for each
remaining task.

\begin{wrapfigure}[11]{r}{0.6\linewidth}
    \vspace{-1.8em}
    \centering
    \begin{minipage}[t]{0.49\linewidth}
        \centering
        \includegraphics[width=\linewidth]{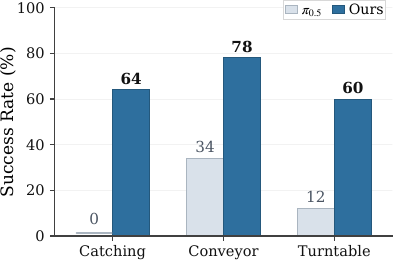}
        \vspace{-1.8em}
        {\scriptsize (a) Success rate}
    \end{minipage}
    \hspace{-0.01\linewidth}
    \begin{minipage}[t]{0.49\linewidth}
        \centering
        \includegraphics[
            width=\linewidth,
            trim=14 6 14 6,
            clip
        ]{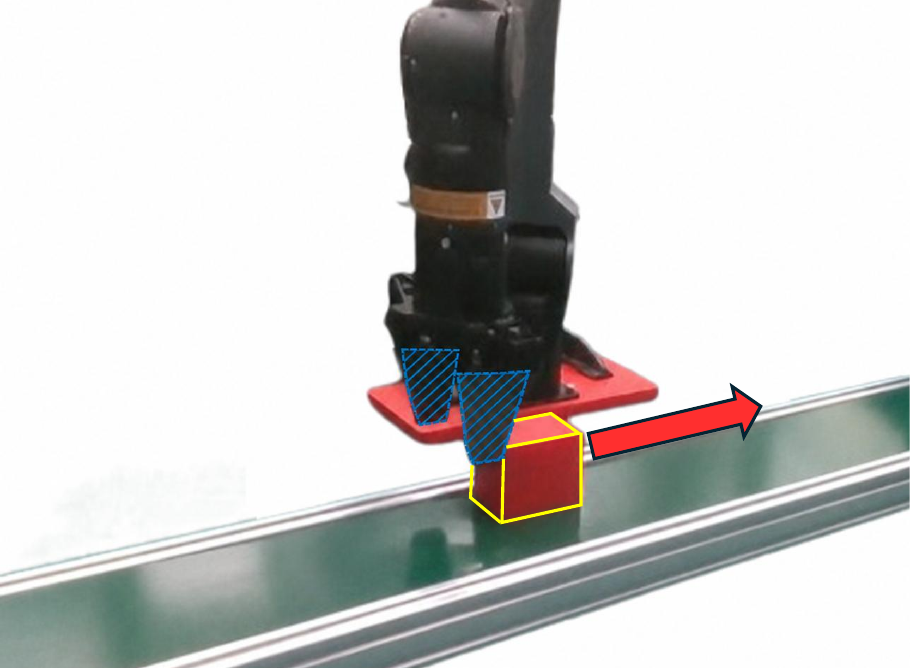}
        \vspace{-1.8em}
        {\scriptsize (b) Reactive failure}
    \end{minipage}
    \vspace{0.8em}
    \caption{
    Real-world success rates across three dynamic grasping tasks and a representative reactive-tracking failure on conveyor interception.
    }
    \label{fig:real_world_results}
    \vspace{-0.8em}
\end{wrapfigure}
Table~\ref{tab:dynamic_main_results} shows that PAF-Grasp achieves the highest success rate across all four tasks. It improves the
average success rate (SR) from $24.5\%$ for the strongest baseline to $62.0\%$, a $37.5$
percentage-point gain. The largest improvements appear on Catch ($11\%$ to
$56\%$) and Turntable ($26\%$ to $62\%$), where successful execution requires
anticipating the object's future position or phase rather than reacting to the
current pose. Our method also achieves the best SR on Conveyor and Ramp,
reaching $78\%$ and $52\%$, respectively. Although some baselines report shorter PL (Path Length) or TCT (Task Completion Time) on specific tasks, these metrics are averaged only over successful trials and should be interpreted with
SR. PAF-Grasp improves success substantially while maintaining comparable PL
and TCT, indicating that the gains come from future interception rather than conservative motion. Observed failures mainly arise from grasp execution errors, the absence of IK-feasible interception candidates, and imperfect initial-velocity estimates.
\begin{table*}[t]
\centering
\small
\setlength{\tabcolsep}{3.5pt}
\caption{
Overall dynamic grasping performance. Ours uses SE(3) forecasting and
reachability-aware contact-time selection. SR is over all trials; PL and TCT are over successful trials only.}
\label{tab:dynamic_main_results}
\resizebox{\textwidth}{!}{
\begin{tabular}{l|ccc|ccc|ccc|ccc|c}
\toprule
\multirow{2}{*}{Method}
& \multicolumn{3}{c|}{Catch}
& \multicolumn{3}{c|}{Conveyor}
& \multicolumn{3}{c|}{Turntable}
& \multicolumn{3}{c|}{Ramp}
& \multirow{2}{*}{Avg. SR $\uparrow$} \\
\cmidrule(lr){2-4}\cmidrule(lr){5-7}\cmidrule(lr){8-10}\cmidrule(lr){11-13}
& SR $\uparrow$ & PL $\downarrow$ & TCT $\downarrow$
& SR $\uparrow$ & PL $\downarrow$ & TCT $\downarrow$
& SR $\uparrow$ & PL $\downarrow$ & TCT $\downarrow$
& SR $\uparrow$ & PL $\downarrow$ & TCT $\downarrow$
&  \\
\midrule
$\pi_0$-FAST~\citep{pertsch2025fastefficientactiontokenization}                         & 8 & 0.16 & 1.32 & 16 & 0.46 & 2.34 & 8 & \textbf{0.33} & \textbf{1.96} & 8 & 0.25 & 3.03 & 10 \\
$\pi_0$~\citep{black2026pi0visionlanguageactionflowmodel}                              & 6 & 0.12 & 0.56 & 50 & 0.44 & \textbf{2.3} & 16 & 0.34 & 2.23 & 26 & 0.18 & 1.99 & 24.5 \\
$\pi_{0.5}$~\citep{intelligence2025pi05visionlanguageactionmodelopenworld}                           & 11 & 0.11 & 0.55 & 46.0 & 0.44 & 2.36 & 16 & \textbf{0.33} & 2.10 & 24 & 0.18 & 2.09 & 24.3\\
VLASH~\citep{Tang2025VLASHRV}                            & 6 & 0.11 & 0.54 & 32 & 0.45 & 2.74 & 26 & 0.43 & 2.4 & 12 & 0.25 & 2.60 & 19 \\
GR00T N1.7~\citep{nvidia2025gr00tn1openfoundation}  & 4  & \textbf{0.06} & \textbf{0.47} & 24 & 0.47 & 2.92 & 12 & 0.46 & 2.43 & 2 & 0.18 & 2.01 & 10.5 \\
Ours      & \textbf{56} & 0.11 & 0.54 & \textbf{78} & \textbf{0.388} & 2.47 & \textbf{62} & 0.51 & 2.32 & \textbf{52} & \textbf{0.13} & \textbf{1.74} & \textbf{62} \\
\bottomrule
\end{tabular}
}
\vspace{-1.3em}
\end{table*}

\textbf{Real-world experiments.}
We evaluate our framework on real-world dynamic grasping tasks using an AgileX
Piper robot arm and an Intel RealSense D435i RGB-D camera. Object-level SE(3)
states are estimated using object meshes and FoundationPose++~\cite{
foundationposeplusplus,FoundationPose,cheng2024putting}. We evaluate three tasks: parabolic ball catching, conveyor-like interception,
and turntable handle grasping. We compare against \(\pi_{0.5}\) with Real-Time Chunking
(RTC)~\citep{black2026realtime}, fine-tuned on 50 successful expert
trajectories per task. We use the same fixed RTC hyperparameters across all tasks without task-specific tuning based on final evaluation results. It receives the same RGB stream,
language instruction, and robot proprioception as our method, but not object
SE(3) histories, future forecasts, or selected contact states. 

We conduct $50$ real-world evaluation trials for each method on each task.
As shown in Fig.~\ref{fig:real_world_results}(a), our framework achieves
\(64\%\) vs. \(0\%\) success on catching, \(78\%\) vs. \(34\%\) on conveyor
interception, and \(60\%\) vs. \(12\%\) on turntable handle grasping.
Fig.~\ref{fig:real_world_results}(b) shows a reactive-tracking failure: when the target velocity exceeds the arm's tracking capability over the latency window, the commanded contact pose becomes stale before execution. This highlights the need for selecting reachable contact-time states from forecasted SE(3) trajectories, rather than reactively chasing stale current poses.

\subsection{Evaluation of Future Prediction Representations}
We investigate what form of future representation is most suitable for
proactive dynamic grasping. In particular, we compare dense visual future
prediction with object-centric state forecasting. While dense predictive world
models forecast future observations in pixel space, our forecaster directly
predicts future SE(3) rigid-body states. This experiment evaluates the
suitability of different future representations for dynamic grasping, rather
than providing an input-matched comparison between general-purpose visual
prediction models and our specialized SE(3) forecaster.

\begin{figure*}[!ht]
    \centering
    \vspace{-0.5em}

    \begin{minipage}[t]{0.30\textwidth}
        \vspace{0pt}
        \centering
        \footnotesize
        \renewcommand{\arraystretch}{0.8}
        \setlength{\tabcolsep}{2.0pt}

        \captionof{table}{
        Contact-relevant future prediction quality.
        }
        \label{tab:video_gen_compare}
        \vspace{-0.5em}

        \resizebox{0.95\linewidth}{!}{
        \begin{tabular}{lcc}
        \toprule
        Method & mIoU $\uparrow$ & Time $\downarrow$ \\
        \midrule
        CogVideoX & 0.0036 & 180 \\
        Wan 2.2   & 0.0225 & 41 \\
        Veo 3.1   & 0.0334 & 71 \\
        Kling 3.0 & 0.0466 & 30 \\
        Ours      & \textbf{0.2762} & \textbf{0.015} \\
        \bottomrule
        \end{tabular}
        }
        \vspace{0.3em}

        \includegraphics[
            width=1.0\linewidth,
            trim=2 2 2 2,
            clip
        ]{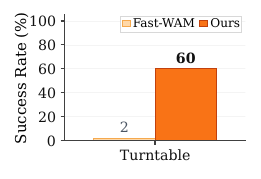}
        \vspace{-2.2em}
        \captionof{figure}{
        Success rate on turntable handle grasping.
        }
        \label{fig:turntable_fastwam}
    \end{minipage}
    \hspace{-0.01\textwidth}
    \begin{minipage}[t]{0.68\textwidth}
        \vspace{0pt}
        \centering
        \setlength{\tabcolsep}{1.5pt}
        \renewcommand{\arraystretch}{0.85}
        \begin{tabular}{cc}
            \includegraphics[width=0.49\linewidth]{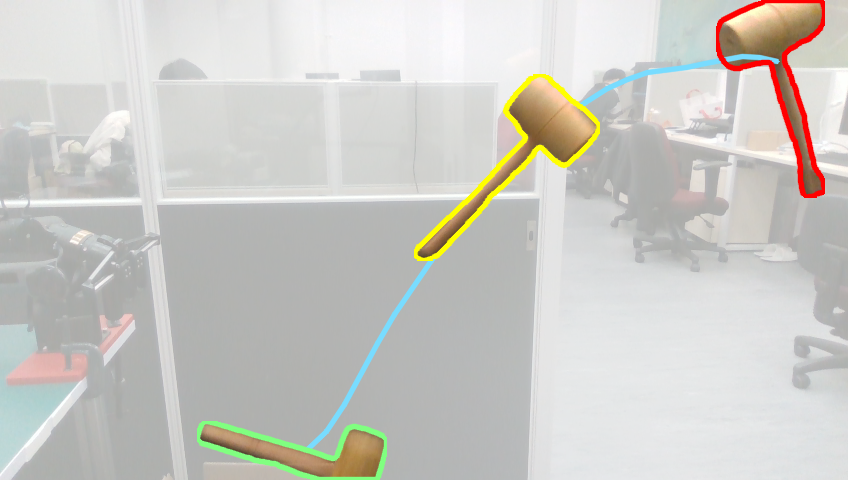} &
            \includegraphics[width=0.49\linewidth]{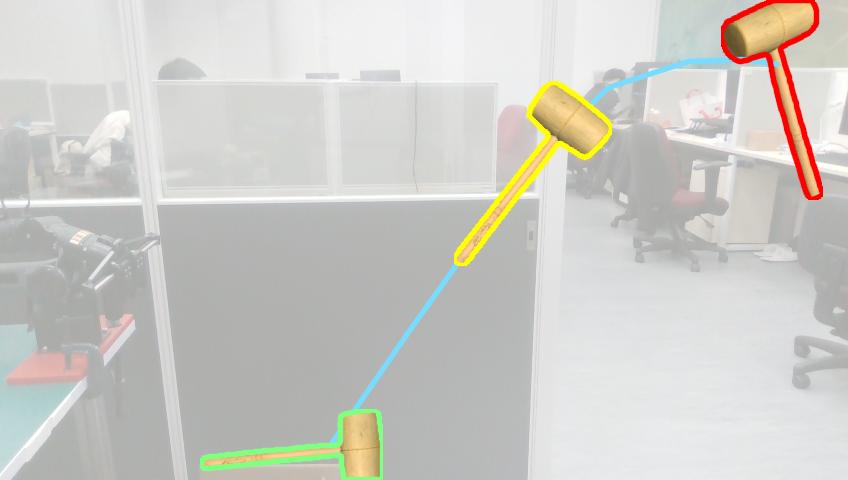} \\
            \scriptsize GT &
            \scriptsize Ours \\
            \includegraphics[width=0.49\linewidth]{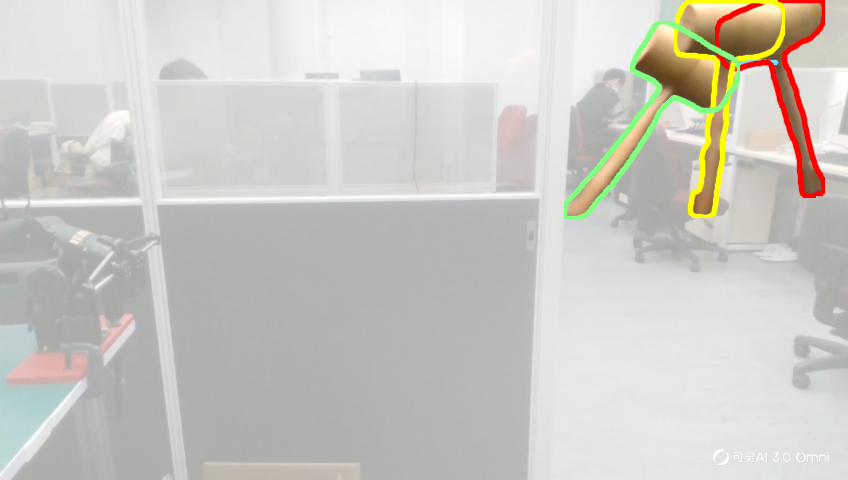} &
            \includegraphics[width=0.49\linewidth]{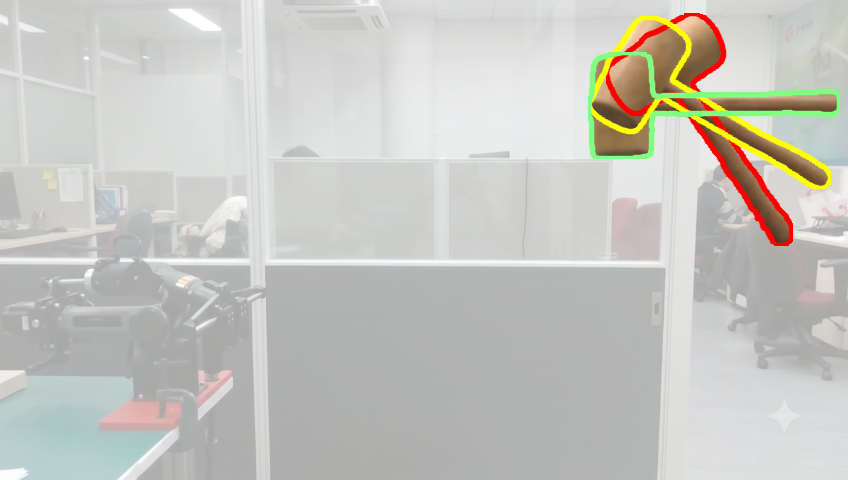} \\
            \scriptsize Kling &
            \scriptsize Veo
        \end{tabular}

        \vspace{-0.4em}
        \captionof{figure}{
        Qualitative comparison of future motion prediction.
        }
        \label{fig:video_generation_compare}
    \end{minipage}

    \vspace{-1.0em}
\end{figure*}

\textbf{Comparison with general video generation models.}
We compare with Wan2.2-TI2V-5B~\citep{wan2025wanopenadvancedlargescale},
CogVideoX1.5-5B-I2V~\citep{yangcogvideox},
Veo 3.1~\citep{wiedemer2025videomodelszeroshotlearners}, and
Kling 3.0 Omni~\citep{klingteam2025klingomnitechnicalreport} under
single-image prompting. We evaluate all methods with the same mask-mIoU protocol: video models generate future clips from a single image and prompt and their target-object masks are extracted using SAM2~\citep{Ravi2024SAM2S}.
Ours renders the predicted SE(3) trajectory from the tracked object mesh into the same camera view. Table~\ref{tab:video_gen_compare} and
Fig.~\ref{fig:video_generation_compare} show that video models can produce visually plausible futures, but their object motion is often physically inconsistent and unsuitable for real-time interception.

To examine whether temporal visual context closes this gap, we additionally
evaluate 3DScenePrompt\citep{lee20263d} using multi-frame RGB observations matched to our
pose-history window. It achieves 0.0584 mIoU, compared with 0.2762 for our SE(3) forecast. In contrast, our SE(3)
forecaster predicts physics-aligned future states at low latency.

\begin{wrapfigure}[11]{r}{0.47\linewidth}
    \vspace{-1.7em}
    \centering
    \includegraphics[
        width=\linewidth,
        trim=2 2 2 2,
        clip
    ]{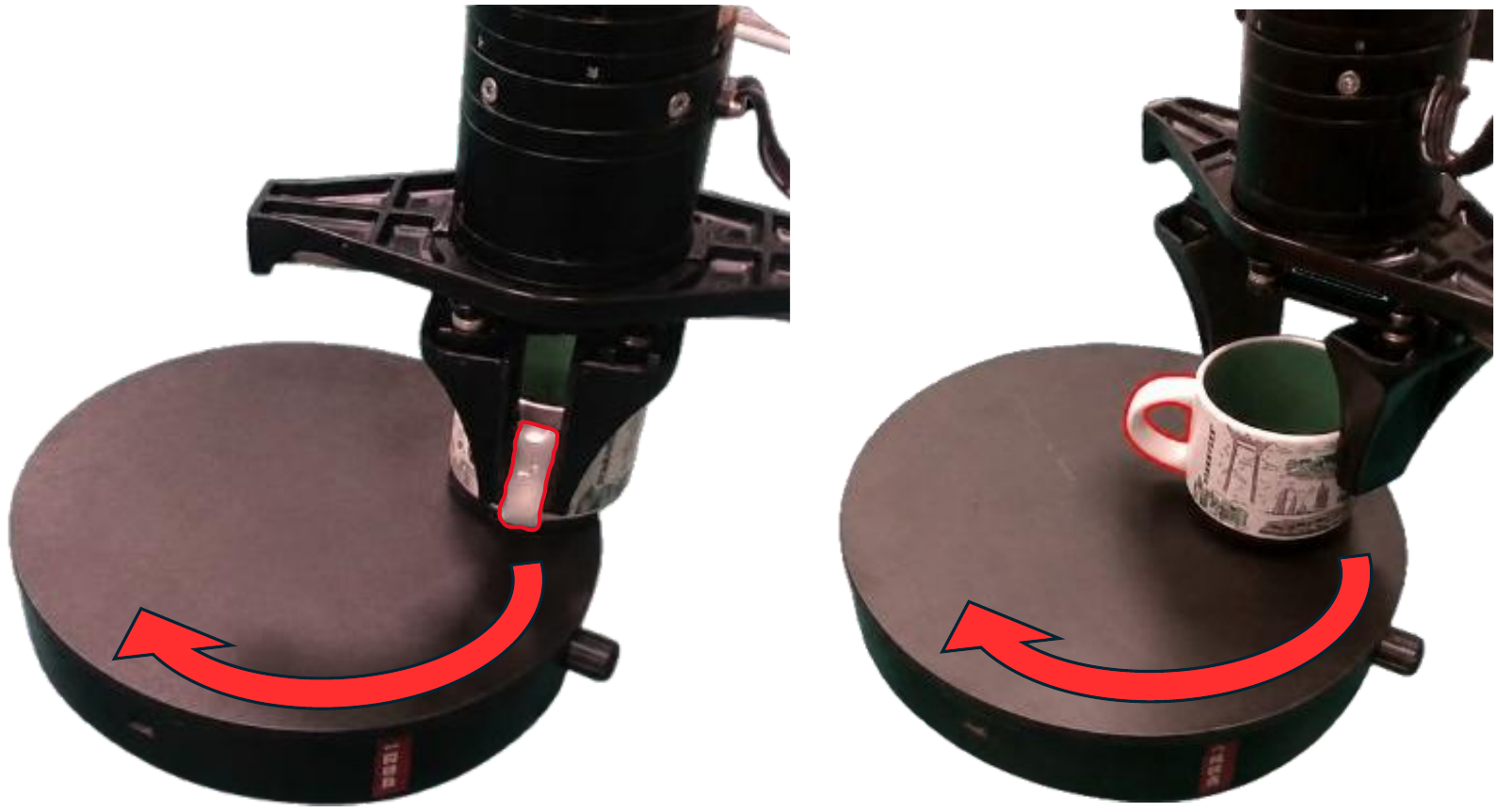}
    \vspace{-1.0em}
    \caption{
    Fast-WAM (right) failure: the gripper reaches the cup but misses the rotating handle.}
    \label{fig:turntable_failure}
\end{wrapfigure}
\textbf{Comparison with world action model.}
The previous comparison evaluates dense visual future prediction as an explicit
future representation. We further ask whether such representations are sufficient
when used inside a real-time action model rather than queried as a separate video
rollout. To this end, we compare with Fast-WAM~\citep{fastwam} on turntable
handle grasping, where success requires matching the future rotational phase of
a rotating cup handle. Fast-WAM uses pretrained Wan2.2-5B~\citep{
wan2025wanopenadvancedlargescale} as its video backbone and avoids explicit
video rollout at inference time through video-action co-training~\citep{
ye2026worldactionmodelszeroshot, li2026causalworldmodelingrobot,
motubrainteam2026motubrainadvancedworldaction}, making it a strong low-latency
world-action baseline. As shown in Fig.~\ref{fig:turntable_fastwam}, Fast-WAM
achieves \(2\%\) success, whereas PAF-Grasp reaches \(60\%\).
Fig.~\ref{fig:turntable_failure} shows the typical failure mode: Fast-WAM can
move the gripper near the cup, but without explicit future SE(3) state
prediction, it does not reliably choose an orientation-aware grasp pose and often
misses the rotating handle. This suggests that visual future priors alone remain insufficient when successful interception depends on metrically accurate contact-time rigid-body state.

\subsection{Why Neural ODE Forecasting?}
Having established SE(3) state as the prediction target, we next study what forecasting model is needed. We first ask whether a learned forecaster is necessary when the motion structure is known. To isolate forecasting from interception, we compare our NODE forecaster with task-aware classical predictors using the same noisy SE(3) pose history, without ground-truth velocity, and the same reachability-aware selector. The classical predictors instantiate the known motion structure of each family: ballistic motion for toss, constant velocity for conveyor, circular motion with a known rotation center for turntable, and constant acceleration inferred from pose history for ramp. Rotational motion is estimated from the same noisy pose history for all methods. 

\begin{wraptable}[13]{r}{0.48\linewidth}
\vspace{-1.2em}
\centering
\caption{Task-aware classical predictors vs.\ NODE under matched noisy
SE(3) inputs and the same selector. Pos-FDE: positional final displacement
error; Rot-FDE: final rotation error.}
\vspace{-0.3em}
\label{tab:classical_node}
\setlength{\tabcolsep}{3pt}
\renewcommand{\arraystretch}{0.95}
\resizebox{\linewidth}{!}{
\begin{tabular}{llccc}
\toprule
Task & Pred. & Pos-FDE $\downarrow$ & Rot-FDE $\downarrow$ & SR $\uparrow$ \\
\midrule
\multirow{2}{*}{Toss}
& Classical & 0.203 & 12.99$^\circ$ & 36\% \\
& NODE      & \textbf{0.202} & \textbf{10.64$^\circ$} & \textbf{56\%} \\
\midrule
\multirow{2}{*}{Conveyor}
& Classical & 0.076 & 10.48$^\circ$ & 78\% \\
& NODE      & \textbf{0.057} & \textbf{0.55$^\circ$} & 78\% \\
\midrule
\multirow{2}{*}{Turntable}
& Classical & \textbf{0.049} & 10.60$^\circ$ & 54\% \\
& NODE      & 0.059 & \textbf{3.94$^\circ$} & \textbf{62\%} \\
\midrule
\multirow{2}{*}{Ramp}
& Classical & 0.505 & 6.24$^\circ$ & \textbf{60\%} \\
& NODE      & \textbf{0.266} & \textbf{0.51$^\circ$} & 52\% \\
\bottomrule
\end{tabular}
}
\end{wraptable}
As shown in Table~\ref{tab:classical_node}, classical predictors remain
competitive when the motion structure is known: they match NODE on
conveyor and achieve a higher success rate on ramp. NODE nevertheless
reduces rotational forecasting error across all four motion families
and improves the overall interception success rate from 57\% to 62\%.
Rather than uniformly outperforming motion-specific analytical models,
the learned forecaster provides a common physics-structured formulation
without requiring a separate rollout law for each motion family. This matched noisy-input evaluation differs from the following forecasting
ablations, which use simulator ground-truth poses to isolate the forecasting
architecture.

Given a learned forecaster, we next ask whether continuous-time dynamics modeling is preferable to recursive sequence prediction. We compare a recursive
autoregressive forecaster (AR) with our continuous-time Neural ODE forecaster
(NODE) to study which temporal model is suitable for this abstraction. Table~\ref{tab:ar_node_main} shows that
AR is competitive on regular short-horizon motions, but NODE performs better on
dynamically richer regimes such as slide and toss, where future motion is
affected by gravity, friction, and rotational inertial effects. Beyond the
supervised horizon, this advantage becomes more pronounced in the per-horizon
error curves. As shown in Fig.~\ref{fig:extrapolation_curves}, AR and NODE are
comparable within the supervised range, but AR errors grow rapidly after horizon
\(H=20\). By \(H=40\), NODE maintains lower error on both conveyor-like position
prediction and turntable rotation prediction. This smoother error growth
indicates that continuous-time dynamics provides better long-horizon stability
for contact-state forecasting than kinematic rollout.

\begin{wraptable}[5]{r}{0.28\linewidth}
    \vspace{-1.2em}
    \centering
    \scriptsize
    \setlength{\tabcolsep}{2.0pt}
    \renewcommand{\arraystretch}{0.82}

    \caption{Toss ablation.}
    \label{tab:inertia_ablation}
    \vspace{-0.5em}

    \resizebox{\linewidth}{!}{
    \begin{tabular}{lcc}
    \toprule
    Method & Pos-FDE $\downarrow$ & Rot-FDE $\downarrow$ \\
    \midrule
    AR
    & 0.044 & 12.693 \\
    NODE w/o $I^B$
    & \textbf{0.023} & 11.241 \\
    NODE w/ $I^B$
    & \textbf{0.023} & \textbf{4.745} \\
    \bottomrule
    \end{tabular}
    }
    \vspace{-1.0em}
\end{wraptable}
We further ablate inertia conditioning on toss forecasting. As shown in
Table~\ref{tab:inertia_ablation}, replacing AR with NODE reduces Pos-FDE from
\(0.044\,\mathrm{m}\) to \(0.023\,\mathrm{m}\), while adding the body-frame
inertia prior \(\psi=I^B\) reduces Rot-FDE from \(11.24^\circ\) to
\(4.75^\circ\). On toss, this shows that inertia conditioning substantially improves rotational forecasting. The recovered inertia prior provides object-specific conditioning for the corresponding motion-family forecaster without object-specific retraining.
\begin{table*}[!ht]
    \centering
    \vspace{-0.2em}

    \begin{minipage}[t]{0.32\textwidth}
        \vspace{0pt}
        \centering
        \scriptsize
        \setlength{\tabcolsep}{1.7pt}
        \renewcommand{\arraystretch}{0.78}

        \captionof{table}{In-range FDE.}
        \label{tab:ar_node_main}
        \vspace{-0.45em}

        \resizebox{\linewidth}{!}{
        \begin{tabular}{llcc}
        \toprule
        Task & Method & Pos-FDE $\downarrow$ & Rot-FDE $\downarrow$ \\
        \midrule
        \multirow{2}{*}{Conveyor}
        & AR   & 0.010 & \textbf{0.082} \\
        & NODE & \textbf{0.010} & 0.225 \\
        \midrule
        \multirow{2}{*}{Slide}
        & AR   & 0.234 & 0.730 \\
        & NODE & \textbf{0.133} & \textbf{0.665} \\
        \midrule
        \multirow{2}{*}{Toss}
        & AR   & 0.047 & 1.099 \\
        & NODE & \textbf{0.023}& \textbf{0.669} \\
        \midrule
        \multirow{2}{*}{Turntable}
        & AR   & \textbf{0.005} & \textbf{1.545} \\
        & NODE & \textbf{0.005} & 1.565 \\
        \bottomrule
        \end{tabular}
        }
    \end{minipage}
    \hspace{0.015\textwidth}
    \begin{minipage}[t]{0.65\textwidth}
        \vspace{0pt}
        \centering
        \begin{minipage}[t]{0.49\linewidth}
            \centering
            \includegraphics[
                width=\linewidth,
                trim=1 1 1 1,
                clip
            ]{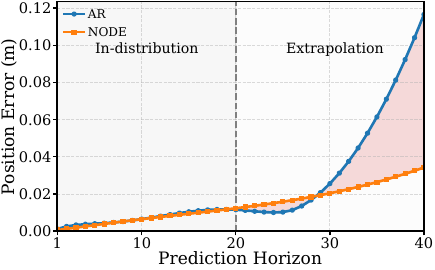}
        \end{minipage}
        \hfill
        \begin{minipage}[t]{0.49\linewidth}
            \centering
            \includegraphics[
                width=\linewidth,
                trim=1 1 1 1,
                clip
            ]{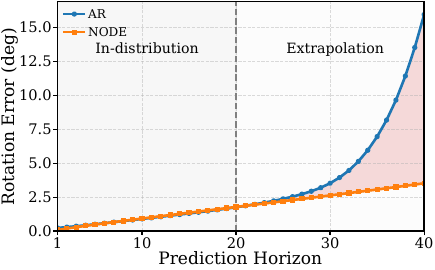}
        \end{minipage}

        \vspace{-0.5em}
        \captionof{figure}{
        Extrapolation beyond the supervised horizon. Left: conveyor position
        error. Right: turntable rotation error.
        }
        \label{fig:extrapolation_curves}
    \end{minipage}

    \vspace{-1.0em}
\end{table*}

\section{Conclusion}
\label{sec:conclusion}
We presented PAF-Grasp, a physics-aware forecast-then-select framework for
real-time proactive dynamic grasping. PAF-Grasp follows a forecast-then-select principle: it forecasts the target object's future SE(3) rigid-body trajectory and then selects a reachable contact-time state under timing and kinematic constraints. This explicit physical state abstraction connects
motion forecasting with robot reachability, enabling the system to act
proactively rather than react to instantaneous observations. Across simulation and real-world experiments, this compact physical abstraction
improves success over reactive visuomotor baselines while remaining compatible
with real-time control. Comparisons with prediction models based on dense visual representations further suggest that visually plausible futures alone are insufficient for real-time dynamic grasping, where success depends on metrically accurate contact-time
object state. Together, these results suggest that proactive dynamic grasping requires a
future abstraction that is physically grounded, time-aware, and directly
actionable for real-time contact decision making.

\textbf{Limitations.}
Our framework currently focuses on short-horizon, continuous single-rigid-body
SE(3) motion. Deformable objects require non-rigid state representations and
deformation-aware dynamics, while coupled interactions, contact transitions,
and collisions require interaction-aware or hybrid dynamics with contact-aware
states or impulse-based corrections. Our inertia prior further assumes a coarse
single-material density. Spatially varying mass distributions remain outside the
current scope. A natural extension is 3D part segmentation with per-part density
estimation and inertia aggregation, or online refinement of physical priors from
observed motion. The forecaster also depends on accurate initial velocity and
RGB-D pose tracking, which can degrade under fast motion blur or depth
corruption. Event-based tracking such as Event6D~\citep{kang2026event6d} may
improve high-speed perception. Finally, performance is constrained by the
workspace and kinematic singularities of the lightweight AgileX Piper arm,
motivating deployment on more capable mobile or legged manipulators.


\clearpage
\acknowledgments{The research presented in this paper was conducted in the JC STEM Lab of Machine Learning and Computer Vision, funded by The Hong Kong Jockey Club Charities Trust. This work was supported in part by the Global STEM Professorship Scheme of the Hong Kong Special Administrative Region and in part by the National Natural Science Foundation of China under Grant No. 62106236.}


\bibliography{reference}  

\clearpage
\appendix
%

\section{Implementation Details}
\label{app:implementation}

\paragraph{SE(3) Forecaster training.}
Table~\ref{tab:forecaster_training_config} summarizes the training configuration
used for the Neural ODE forecaster. The temporal resolution is \(\Delta t=0.05\,\mathrm{s}\) after downsampling. We use task-specific observation and prediction horizons according to the characteristic duration of each motion type, as summarized in Table~\ref{tab:task_horizons}. RGB and depth inputs are not used; the model is trained from SE(3) pose histories and velocity information.

\begin{table}[h]
\centering
\small
\caption{Task-specific observation and prediction horizons.}
\label{tab:task_horizons}
\begin{tabular}{lccc}
\toprule
Task & Motion type & Observation horizon \(K\) & Prediction horizon \(H\) \\
\midrule
Conveyor & Linear & \(4\) & \(20\) \\
Turntable & Rotation & \(4\) & \(20\) \\
Ramp sliding & Sliding & \(4\) & \(16\) \\
Parabolic catching & Projectile & \(4\) & \(12\) \\
\bottomrule
\end{tabular}
\end{table}

\begin{table}[t]
\centering
\small
\caption{Training configuration for the Neural SE(3) forecaster.}
\label{tab:forecaster_training_config}
\begin{tabular}{ll}
\toprule
Component & Setting \\
\midrule
ODE solver & Dopri5, rtol \(=10^{-5}\), atol \(=10^{-6}\) \\
Context / dynamics dim. & \(32 / 64\) \\
MLP hidden dims. & \([64,128,128,64]\) \\
Velocity estimator & GRU, hidden dim. \(64\), 1 layer, dropout \(0.1\) \\
Optimization & AdamW, lr \(10^{-4}\), wd \(10^{-5}\), cosine decay \\
Batch size / epochs & \(512 / 3000\) \\
Gradient clipping & \(1.0\) \\
\bottomrule
\end{tabular}
\end{table}
The model is trained with
\[
\mathcal{L}
=
\lambda_{pos}\mathcal{L}_{pos}
+
\lambda_{rot}\mathcal{L}_{rot}
+
\lambda_{vel}\mathcal{L}_{vel},
\]
where \(\mathcal{L}_{pos}\) and \(\mathcal{L}_{vel}\) are MSE losses and
\(\mathcal{L}_{rot}\) is a sign-invariant quaternion chordal loss. We use
\((\lambda_{pos},\lambda_{rot},\lambda_{vel})=(1,1,0.05)\).

\begin{table}[t]
\centering
\small
\caption{Configuration for the autoregressive Transformer forecaster baseline.}
\label{tab:transformer_config}
\begin{tabular}{ll}
\toprule
Component & Setting \\
\midrule
Input representation & SE(3) pose history; no RGB/depth/velocity \\
Transformer dim. & \(d_{\mathrm{model}}=128\) \\
Attention heads & \(8\) \\
Encoder / decoder layers & \(4 / 4\) \\
Feedforward dim. & \(512\) \\
Dropout & \(0.1\) \\
Optimization & AdamW, lr \(10^{-5}\), wd \(10^{-5}\), cosine decay \\
Batch size / epochs & \(512 / 3000\) \\
Loss & position MSE + chordal quaternion loss \\
Gradient clipping & \(1.0\) \\
\bottomrule
\end{tabular}
\end{table}

For the autoregressive baseline used in Sec.~4.3, we train a Transformer encoder--decoder that predicts future SE(3) poses directly from pose histories. Its configuration is summarized in Table~\ref{tab:transformer_config}.

\section{Simulation Training and Task Settings}
\label{app:sim_tasks}
\subsection{SE(3) Forecaster Training}
We train the SE(3) forecaster using object-level rigid-body trajectories
generated in Genesis. The training set contains randomized primitive objects,
including spheres, cuboids, and cylinders, and each episode records the object's
SE(3) pose, linear velocity, angular velocity, and physical parameters such as
object scale, density, friction, and body-frame inertia. The initialization and
motion parameters are randomized separately for each motion type, so that the
forecaster observes diverse rigid-body dynamics rather than trajectories from a
single object or task instance. We generate \(1000\) trajectories for each
motion type, including parabolic motion, conveyor-like transport, turntable
rotation, and ramp sliding. These trajectories are downsampled to
\(\Delta t=0.05\,\mathrm{s}\) and converted into sliding-window training samples
according to the task-specific observation and prediction horizons in
Table~\ref{tab:task_horizons}.

\paragraph{Parabolic motion.}
For projectile motion, we generate toss trajectories using rigid primitive
objects, including boxes and spheres. Unless otherwise specified, the total
initial speed is uniformly sampled from \([1.0,7.0]\,\mathrm{m/s}\), and the
initial release height is sampled from \([0.5,2.5]\,\mathrm{m}\). The horizontal azimuth is uniformly sampled from \([0,2\pi)\), and the elevation angle is sampled from \([\pi/9,\pi/2]\), producing upward projectile motions under gravity. For box objects, we additionally sample a random rotation axis and an angular speed from \([2.0,8.0]\,\mathrm{rad/s}\) to induce rigid-body rotation during flight. The simulator uses a quadratic drag coefficient of \(0.01\). To emulate pose-estimation noise, we perturb the observed trajectory with Gaussian translation noise of standard deviation \(0.02\,\mathrm{m}\) and rotation noise of standard deviation \(1.0^\circ\).

\paragraph{Conveyor motion.}
For conveyor-like transport, we simulate constant-velocity planar object
motion. For each episode, the object's transport speed is sampled from
\([0.04,0.40]\,\mathrm{m/s}\), and the motion direction is sampled uniformly
from \([0,2\pi)\), producing a horizontal velocity \(v=(v_x,v_y,0)\). We add
Gaussian translation noise with standard deviation \(0.005\,\mathrm{m}\) and
rotation noise with standard deviation \(1.0^\circ\) to emulate noisy SE(3)
tracking. This task is dynamically simpler than projectile motion, but still
requires contact-time prediction: the robot must move to where the object will
be when the end-effector arrives, rather than reacting to the currently observed
pose.

\paragraph{Turntable motion.}
For rotational motion, we randomize the object's initial phase, radial distance
to the turntable center, angular velocity, and object pose. The angular velocity is sampled from \([0.4,1.0]\,\mathrm{rad/s}\), the object-center distance from the rotation center is sampled from \([0.1,0.4]\,\mathrm{m}\), and the initial phase is uniformly sampled from \([0,2\pi)\). To emulate pose-estimation noise, we perturb the observed SE(3) trajectory with Gaussian translation noise of standard deviation \(0.005\,\mathrm{m}\) and rotation noise of standard deviation \(1.0^\circ\). This setting requires forecasting both the future object position and the rotational phase of the task-relevant contact frame. 

\paragraph{Ramp sliding.}
For ramp sliding, we simulate objects sliding down an inclined plane. For each
episode, the ramp angle is sampled from \([10^\circ,45^\circ]\), and the
friction coefficient is sampled from \([0.1,0.4]\). The resulting motion is
non-uniform due to gravity and friction, making this setting more challenging
than constant-velocity conveyor motion. To emulate noisy SE(3) tracking, we add
Gaussian translation noise with standard deviation \(0.01\,\mathrm{m}\) and
rotation noise with standard deviation \(1.0^\circ\). This task evaluates
whether the forecaster can predict accelerating object motion along a constrained surface.

\subsection{Simulation Task Settings}
\label{app:sim_task_settings}
All simulation tasks follow the same forecast-guided contact-state selection
principle. At each control step, the SE(3) forecaster predicts a future object
trajectory from the recent pose history. The predicted trajectory is mapped to a sequence of task-relevant contact frames, and the reachability-aware selector
filters candidates using workspace, latency, and timing constraints.

We use different execution modes for slow and fast dynamic tasks. For slower
tasks, including conveyor interception, turntable grasping, and ramp sliding,
the feasible contact window is usually long enough to support a two-stage
execution. The robot first moves to an intermediate pre-contact hover pose
associated with a selected future contact state. As the object approaches the
contact window, the selector is updated with the latest forecast and the robot
executes the final interception or grasp at the predicted contact-time SE(3)
state.

For fast projectile catching, we do not use a separate hover stage. Since the
available contact window is short, the selector directly chooses the earliest
reachable safe candidate along the predicted trajectory, and the robot executes
a fast interception motion toward that contact-time state. This matches the
fast-task selection rule in the main paper, where fast motions prioritize
minimum estimated reach time, while slower tasks select a desired contact time
within the feasible window.

\paragraph{Basket catching.}
For the basket-catching task, the robot first grasps the handle of a basket and
keeps the gripper closed throughout the catching phase. A ball is then tossed
toward the workspace with randomized initial position and velocity. The ball
starts from \(x\in[1.3,1.5]\,\mathrm{m}\), \(y\in[-0.2,0.2]\,\mathrm{m}\),
and \(z\in[0.5,1.0]\,\mathrm{m}\). Its initial speed is sampled from
\([2.0,2.5]\,\mathrm{m/s}\), the elevation angle from
\([\pi/5,\pi/4]\), and the azimuth around the direction toward the robot. The
robot must move the basket to intercept the ball and catch it inside the basket
opening. The robot observes the scene from two fixed external cameras, and each
episode is successful if the ball remains inside the basket after contact.

\paragraph{Conveyor interception.}
For the conveyor task, a cube is placed on a moving belt and transported toward
the robot workspace. At the beginning of each episode, we randomize the belt
speed, the belt starting position, and the cube's lateral offset. The belt speed is sampled from \([0.08,0.20]\,\mathrm{m/s}\), the initial belt position is sampled along the transport direction, and the cube offset is sampled from
\([-0.05,0.05]\,\mathrm{m}\) around the nominal interception line. The robot is
initialized from a fixed pre-grasp configuration with the gripper open.

\paragraph{Turntable grasping.}
For the turntable task, a small cylinder is placed on a rotating turntable. At the beginning of each episode, we randomize the cylinder's initial angular phase, radial distance from the turntable center, and turntable angular velocity. The
angular velocity is sampled from \([0.4,1.0]\,\mathrm{rad/s}\), with the rotation direction randomly flipped. The object-center distance from the rotation center is sampled from \([0.1,0.4]\,\mathrm{m}\). The robot is initialized from a fixed pre-grasp configuration with the gripper open.

\paragraph{Ramp sliding.}
For the ramp task, a block is placed near the top of an inclined plane and slides down toward the robot workspace. At the beginning of each episode, we randomize the ramp angle, friction coefficient, and the block's lateral offset. The ramp angle is sampled from \([8^\circ,20^\circ]\), the friction coefficient from \([0.20,0.35]\), and the lateral block offset from
\([-0.12,0.12]\,\mathrm{m}\). We discard configurations where the block does
not slide or where the predicted arrival point falls outside the reachable
interception region. The robot is initialized from a fixed home configuration
with the gripper open.

\section{Real-world Task Settings}
\label{app:real_tasks}

\begin{figure}[t]
    \centering
    \includegraphics[
        width=0.75\linewidth,
    ]{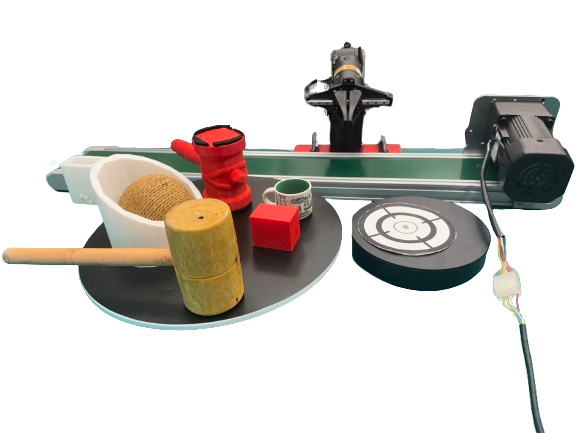}
    \caption{
    Real-world experimental setup for dynamic grasping.
    }
    \label{fig:real_world_setup}
\end{figure}

\paragraph{Experimental setup.}
We conduct real-world experiments on a tabletop dynamic-grasping platform, as
shown in Fig.~\ref{fig:real_world_setup}. The setup consists of an AgileX Piper
robot arm, an Intel RealSense D435i RGB-D camera, a linear conveyor platform, a
rotating turntable, and task objects with different geometries and contact
frames. The robot controller runs at \(100\,\mathrm{Hz}\). For object-centric SE(3) tracking in conveyor and turntable tasks, we use
FoundationPose++~\citep{foundationposeplusplus}, which runs at approximately
\(20\,\mathrm{Hz}\). For sphere catching, where orientation is not
task-relevant, we track the ball mask with Cutie~\citep{cheng2024putting} and recover its 3D center from
RGB-D depth. The forecaster predicts future contact-relevant object states, and the selector converts them into reachable end-effector targets under latency, workspace, and timing constraints. We conduct \(50\) trials per method per real-world task.

\paragraph{Basket-based parabolic catching.}
For real-world parabolic catching, we use a basket-based setup to evaluate
high-speed contact-time interception. The robot first grasps the handle of a
3D-printed basket and keeps the gripper closed during catching. A ball is then
tossed toward the workspace, and the robot moves the basket to intercept the
predicted future ball state. Since the tossed object is a sphere, its orientation
is not task-relevant. We track the ball with Cutie~\citep{cheng2024putting} and recover its 3D center
from RGB-D depth, using the resulting position history as the forecasting input.
The task-relevant contact frame is defined at the basket opening center, and the
selector aligns this frame with the predicted future ball center at the selected
contact time. A trial is counted as successful if the ball enters and remains
inside the basket after contact.

We additionally implemented an orientation-sensitive hammer-catching variant, in
which the robot must make the hammer head strike a target center on the robot
arm. This variant tests a stricter SE(3) contact-frame requirement: successful
contact depends not only on the future 3D position of the hammer, but also on
the future orientation of the task-relevant hammer-head frame. Thus, unlike
basket-based ball catching, which mainly evaluates translational contact-time
prediction, this variant requires aligning a full future SE(3) contact state.

In our real-world setup, high-speed 6D tracking~\citep{foundationposeplusplus} of the asymmetric hammer was
sensitive to motion blur and rapid pose changes, making the measured success
rate strongly coupled to perception stability rather than only to the
interception policy. We therefore use basket-based ball catching as the main
real-world catching benchmark, and include the hammer variant as a supplementary
diagnostic experiment illustrating the importance of full SE(3) future
contact-state prediction.

\paragraph{Conveyor interception.}
For the real-world conveyor task, an object is placed on a linear conveyor belt
and transported toward the robot workspace. The conveyor speed is set between
\(5\) and \(10\,\mathrm{m/min}\), corresponding to approximately
\([0.083,0.167]\,\mathrm{m/s}\). The object's initial position on the belt is
randomized across trials. The object's 6D pose is tracked online with
FoundationPose++~\citep{foundationposeplusplus}. Because the object continues
moving during execution, the selected grasp target must correspond to a future
reachable contact state rather than the current tracked pose. A trial is counted
as successful if the robot grasps and lifts the object from the conveyor.

\paragraph{Turntable interception.}
For the real-world turntable task, a cup-like object is placed on a rotating
turntable, and the robot must grasp a contact frame on the handle region. At the
beginning of each episode, the object's radial distance from the turntable
center is sampled from \([0.10,0.15]\,\mathrm{m}\), and the turntable angular
velocity is sampled from \([0.4,0.6]\,\mathrm{rad/s}\). The object's 6D pose is
tracked online with FoundationPose++~\citep{foundationposeplusplus}. Since the
handle pose changes with the turntable phase, successful interception requires
predicting both the future position and rotational phase of the object. A trial
is counted as successful if the robot grasps the handle and lifts the object from the
turntable.

\paragraph{Evaluation metrics.}
We report success rate (SR) over all evaluation trials. Path length (PL) is
computed as the accumulated end-effector translation during each successful
trial, and task completion time (TCT) is measured from the start of policy
execution to the first detected success. PL and TCT are averaged only over
successful trials.

\section{Interception Selector Details}
\label{app:selector}

We report the hyperparameters used by the reachability-aware interception
selector. The translational and rotational speed bounds \(\bar v_e\) and
\(\bar\omega_e\) are estimated empirically by commanding predefined end-effector
trajectories and measuring the average execution time. We set the system latency
term to \(T_{\mathrm{lat}}=0.1\,\mathrm{s}\), accounting for perception,
forecasting, communication, and controller delay. This value is estimated from
the observed object-motion update interval in our real-world setup: the tracked
object typically moves by approximately \(5\,\mathrm{mm}\) between effective
pose updates, corresponding to about \(100\,\mathrm{ms}\) under the tested motion
ranges. We therefore use \(T_{\mathrm{lat}}=0.1\,\mathrm{s}\) as a conservative
latency compensation term in the selector.

For fast tasks such as parabolic catching, the selector chooses the feasible
candidate with the smallest estimated reach time. We use a small timing margin
\(T_{\mathrm{margin}}=0.2\,\mathrm{s}\) to avoid overly conservative rejection
of near-term interception candidates. For slow tasks such as conveyor
interception, turntable grasping, and ramp sliding, feasible candidates are
sorted by contact time and selected according to the normalized rank parameter
\(\alpha\), with a larger timing margin \(T_{\mathrm{margin}}=0.5\,\mathrm{s}\)
to account for slower execution and pose-tracking uncertainty. We set
\(\alpha=0.0\) for slow tasks, selecting the earliest candidate in the feasible
contact window unless otherwise specified. If no candidate satisfies the
reachability and timing constraints, the robot uses a conservative fallback
target or holds the previous command.

\section{Baseline and Evaluation Protocols}
\label{app:baselines}

We use matched evaluation protocols for simulation, real-world, video-generation,
and world-action baselines. In simulation, all reactive visuomotor baselines are
fine-tuned and evaluated under the same Genesis-generated task data,
initialization distribution, and success criteria. In real-world experiments, we
compare against a strong \(\pi_{0.5}\) baseline with Real-Time Action Chunking~\citep{black2026realtime},
trained on successful expert trajectories collected by PAF-Grasp. For the
simulation and real-world policy baselines, the comparison evaluates the effect
of making reachable future contact states explicit, rather than equalizing
intermediate representations across methods.

\paragraph{Simulation baselines.}
For simulation experiments, we fine-tune \(\pi_0\)~\citep{
black2026pi0visionlanguageactionflowmodel}, \(\pi_{0.5}\)~\citep{
intelligence2025pi05visionlanguageactionmodelopenworld}, \(\pi_0\)-FAST~\citep{
pertsch2025fastefficientactiontokenization}, VLASH~\citep{Tang2025VLASHRV}, and
GR00T N1.7~\citep{nvidia2025gr00tn1openfoundation} using the same training pipeline. All models are fine-tuned with LoRA using a batch size of
\(32\) for \(30{,}000\) training steps on two NVIDIA RTX 5090 GPUs with
\(32\,\mathrm{GB}\) memory each. They are trained on the same Genesis-generated
task data and evaluated under the same initialization distribution and success
criteria.

\paragraph{Real-world baseline.}
For real-world experiments, we use \(\pi_{0.5}\) with Real-Time Action
Chunking~\citep{black2026realtime} as the reactive visuomotor baseline. It is
fine-tuned with the same JAX-based training pipeline on \(50\) successful expert
trajectories collected by PAF-Grasp. The baseline receives the same RGB stream,
language instruction, and robot proprioception as our method, but does not
receive object SE(3) histories, future forecasts, or selected contact states at
test time.

\paragraph{Video generation baselines.}
For video-generation baselines, we use the first RGB frame of each evaluation
sequence as the image prompt and provide a task-specific text prompt describing
the expected future motion. All models are asked to generate a \(1\)-second
future video from a fixed camera view. For the hammer-toss task, we use the
following prompt:
\begin{quote}
\small
Starting from the given input image, generate a realistic fixed-camera
\(1\)-second video. A hammer moves through the air along a smooth parabolic
trajectory under gravity, rotating naturally as a rigid body during flight. The
motion should resemble real projectile motion: the hammer rises slightly,
reaches an arc, then descends and lands within the video duration.
\end{quote}

\begin{wraptable}[9]{r}{0.34\textwidth}
\vspace{-1.0em}
\centering
\footnotesize
\setlength{\tabcolsep}{2.2pt}
\renewcommand{\arraystretch}{0.78}
\caption{
Contact-relevant future prediction quality.
}
\label{tab:app_video_gen_compare}
\vspace{-0.7em}
\resizebox{\linewidth}{!}{
\begin{tabular}{lcc}
\toprule
Method & mIoU $\uparrow$ & Time $\downarrow$ \\
\midrule
CogVideoX & 0.0036 & 180 s \\
Wan 2.2   & 0.0225 & 41 s \\
Veo 3.1   & 0.0334 & 71 s \\
Kling 3.0 & 0.0466 & 30 s \\
\midrule
Ours      & \textbf{0.2762} & \textbf{15 ms} \\
\bottomrule
\end{tabular}
}
\vspace{-1.2em}
\end{wraptable}

For PAF-Grasp, we use the tracked object pose history from
FoundationPose++~\citep{foundationposeplusplus} as input to the SE(3)
forecaster, and render the predicted future poses with the tracked object mesh
to obtain predicted masks. This allows dense visual future prediction and SE(3)
rigid-body forecasting to be evaluated under the same object-centric mask-mIoU
protocol.

Table~\ref{tab:app_video_gen_compare} summarizes the Sec.~4.2 results comparing
future representations for dynamic grasping. Dense video-generation models
predict future appearance, but their object motion is not sufficiently
contact-accurate under the mask-mIoU protocol. PAF-Grasp instead predicts the
future object-centric SE(3) state, achieving higher contact-relevant accuracy
while running in \textbf{15 ms}. This enables real-time contact-state selection
and shows that compact SE(3) futures are better aligned with dynamic grasping
than dense visual futures. The reported \(15\,\mathrm{ms}\) runtime is measured for one forecasting call, including history encoding, velocity estimation, and ODE rollout, but excluding
external perception.

\paragraph{World-action model baseline.}
For the world-action baseline, we fine-tune Fast-WAM on the turntable handle grasping task. Fast-WAM uses Wan2.2-TI2V-5B~\citep{
wan2025wanopenadvancedlargescale} as its video prediction backbone to provide visual future priors during training, and learns an action model through video-action co-training. Unlike imagine-then-execute baselines~\citep{ye2026worldactionmodelszeroshot,
motubrainteam2026motubrainadvancedworldaction,
li2026causalworldmodelingrobot}, Fast-WAM does not perform explicit video rollout at test time. We train Fast-WAM on \(50\) expert trajectories collected by PAF-Grasp, using a batch size of \(32\) for \(5{,}000\) training steps on four
NVIDIA RTX A6000 GPUs with \(48\,\mathrm{GB}\) memory each.

\section{Sim-to-real Pipeline}
\label{app:sim2real}
For each real object, we recover the quantities required by the forecaster from
an RGB-D observation. We first segment the target object using
SAM2~\citep{Ravi2024SAM2S} and reconstruct an object mesh with
TRELLIS-1~\citep{trellis}. Since the generated mesh is scale-normalized, we
recover its metric scale and camera-frame pose using the coarse-to-fine
alignment procedure of OnePoseViaGen~\citep{gengone}. To obtain a coarse
physical prior, we use Qwen3-VL-30B-A3B-Instruct~\citep{bai2025qwen3} to infer
the object's material category from the RGB observation and map it to an
approximate density. Combined with the metric mesh, this yields an approximate
mass and body-frame inertia prior. During deployment, FoundationPose++~\citep{
foundationposeplusplus} tracks the object and provides the past SE(3) state
history, while the recovered geometry and inertia prior are provided to the
forecaster.

\end{document}